\documentclass[10pt,twocolumn]{article}

\usepackage[a4paper,margin=1.8cm]{geometry}
\usepackage[T1]{fontenc}
\usepackage{lmodern}
\usepackage{microtype}

\usepackage{amsmath,amssymb}
\usepackage{graphicx}
\usepackage{booktabs}
\usepackage{float}
\usepackage{stfloats}  
\usepackage{tikz}
\usetikzlibrary{arrows.meta, calc}
\usepackage[font=small]{caption}

\definecolor{oiBlue}{RGB}{0,114,178}
\definecolor{oiOrange}{RGB}{230,159,0}
\definecolor{oiGreen}{RGB}{0,158,115}
\definecolor{oiVermillion}{RGB}{213,94,0}
\definecolor{panelBg}{RGB}{250,250,250}
\definecolor{panelBdr}{RGB}{218,218,218}
\definecolor{hd}{RGB}{40,40,40}
\definecolor{bd}{RGB}{60,60,60}
\definecolor{dm}{RGB}{145,145,145}
\definecolor{ag}{RGB}{115,115,115}
\definecolor{rowG}{RGB}{237,250,244}
\definecolor{rowR}{RGB}{253,241,237}
\definecolor{pillG}{RGB}{225,245,235}
\definecolor{pillGb}{RGB}{0,158,115}
\definecolor{fdBg}{RGB}{244,244,244}
\definecolor{fdBdr}{RGB}{198,198,198}
\definecolor{fdTx}{RGB}{178,178,178}
\definecolor{guarBg}{RGB}{242,251,246}
\definecolor{guarBdr}{RGB}{165,212,188}
\definecolor{thr}{RGB}{180,80,40}
\newcommand{\vp}[4]{%
  \fill[#4!8, draw=#4!30, rounded corners=2pt, line width=0.25pt]
    (#1-0.34,#2-0.10) rectangle (#1+0.34,#2+0.10);
  \node[font=\fontsize{4.5}{5.5}\selectfont\bfseries, text=#4!75!black] at (#1,#2) {#3};
}

\usepackage{enumitem}
\setlist[itemize]{topsep=2pt,itemsep=1pt,parsep=0pt,partopsep=0pt,leftmargin=*}
\setlist[enumerate]{topsep=2pt,itemsep=1pt,parsep=0pt,partopsep=0pt,leftmargin=*}

\usepackage{titlesec}
\titlespacing*{\section}{0pt}{0.9ex plus 0.2ex minus 0.2ex}{0.5ex}
\titlespacing*{\subsection}{0pt}{0.7ex plus 0.2ex minus 0.2ex}{0.35ex}

\usepackage{algorithm}
\usepackage{algpseudocode}

\usepackage{amsthm}
\newtheorem{theorem}{Theorem}
\newtheorem{proposition}{Proposition}
\newtheorem{definition}{Definition}

\usepackage[hidelinks]{hyperref}

\usepackage[numbers,sort&compress]{natbib}

\title{Verifiable Semantics for Agent-to-Agent Communication\thanks{We thank Connie Hsueh, Matt Nour, Ben Bariach, Maartje Nugteren, Deborah Morgan, Michael Bhaskar, and Bea Costa Gomes for helpful comments on this paper.}}
\author{
  Philipp Schoenegger$^{1}$ \quad Matt Carlson$^{2}$ \quad Chris Schneider$^{1}$ \quad Chris Daly$^{1}$ \\[1ex]
  \textnormal{$^{1}$Microsoft AI \quad\quad $^{2}$Wabash College}
}
\date{}

\begin{document}
\maketitle

\begin{abstract}
\textit{Multi-agent Artificial Intelligence (AI) systems require consistent communication, but we lack methods to verify that agents share the same understanding of the terms used. Natural language is interpretable but vulnerable to semantic drift, while learned protocols are efficient but opaque. We propose a certification protocol based on the stimulus-meaning model, where agents are tested on shared observable events and terms are certified if empirical disagreement falls below a statistical threshold. Agents restricting their reasoning to certified terms (``core-guarded reasoning'') achieve provably bounded disagreement. The protocol includes mechanisms for detecting drift (recertification) and recovering shared vocabulary (renegotiation). In simulations with varying degrees of semantic divergence, core-guarding reduces disagreement by 72--96\%. We validate the approach on fine-tuned language models in a content moderation task, achieving 51\% disagreement reduction. Our framework provides a foundation for verifiable agent-to-agent communication.}
\end{abstract}

\begin{figure*}[b]
\centering
\resizebox{\textwidth}{!}{%
\begin{tikzpicture}[>=Stealth, every node/.style={font=\footnotesize}]

\def\pw{5.4}
\def\ph{5.4}
\def\gap{0.85}
\pgfmathsetmacro{\bx}{\pw+\gap}
\pgfmathsetmacro{\cx}{2*(\pw+\gap)}

\foreach \px in {0, \bx, \cx}{
  \fill[panelBg, rounded corners=4pt, draw=panelBdr, line width=0.4pt]
    (\px,0) rectangle (\px+\pw, \ph);
}

\foreach \s/\d in {0/\bx, \bx/\cx}{
  \draw[-{Stealth[length=4pt,width=3pt]}, ag, line width=1.2pt]
    (\s+\pw+0.10, \ph/2) -- (\d-0.10, \ph/2);
}


\node[font=\small\bfseries, text=hd, anchor=west]
  at (0.30, \ph-0.35) {(a)~Stimulus Testing};

\node[font=\fontsize{6.5}{8}\selectfont\itshape, text=bd, anchor=west]
  at (0.30, \ph-0.80) {``Does term $T$ apply to event $e$?''};

\def\hY{4.20}
\def\cE{1.50}
\def\cA{3.10}
\def\cB{4.05}
\def\cR{4.90}

\node[font=\fontsize{5}{6}\selectfont\bfseries, text=dm] at (\cE, \hY) {Event};
\node[font=\fontsize{5}{6}\selectfont\bfseries, text=oiBlue!80] at (\cA, \hY) {$A_1$};
\node[font=\fontsize{5}{6}\selectfont\bfseries, text=oiOrange!80] at (\cB, \hY) {$A_2$};
\draw[panelBdr, line width=0.3pt] (0.20, \hY-0.13) -- (\pw-0.20, \hY-0.13);

\def\ra{3.78}
\fill[rowG, rounded corners=2pt] (0.20,\ra-0.18) rectangle (\pw-0.20,\ra+0.18);
\node[font=\fontsize{5}{6}\selectfont, text=bd] at (\cE,\ra) {$e_1$};
\vp{\cA}{\ra}{assent}{oiBlue}
\vp{\cB}{\ra}{assent}{oiOrange}
\node[font=\fontsize{6}{7}\selectfont\bfseries, text=oiGreen!80] at (\cR,\ra) {\checkmark};

\def\rb{3.32}
\fill[rowG, rounded corners=2pt] (0.20,\rb-0.18) rectangle (\pw-0.20,\rb+0.18);
\node[font=\fontsize{5}{6}\selectfont, text=bd] at (\cE,\rb) {$e_2$};
\vp{\cA}{\rb}{dissent}{oiBlue}
\vp{\cB}{\rb}{dissent}{oiOrange}
\node[font=\fontsize{6}{7}\selectfont\bfseries, text=oiGreen!80] at (\cR,\rb) {\checkmark};

\def\rc{2.86}
\fill[rowR, rounded corners=2pt] (0.20,\rc-0.18) rectangle (\pw-0.20,\rc+0.18);
\node[font=\fontsize{5}{6}\selectfont, text=bd] at (\cE,\rc) {$e_3$};
\vp{\cA}{\rc}{assent}{oiBlue}
\vp{\cB}{\rc}{dissent}{oiOrange}
\node[font=\fontsize{6}{7}\selectfont\bfseries, text=oiVermillion!80] at (\cR,\rc) {$\times$};

\node[font=\small, text=dm] at (\pw/2, 2.35) {$\vdots$};

\node[font=\fontsize{5}{6}\selectfont, text=dm] at (\pw/2, 1.90)
  {Repeat per term $T \in V$};

\draw[panelBdr, line width=0.3pt] (0.20, 1.45) -- (\pw-0.20, 1.45);
\node[font=\fontsize{5}{6}\selectfont, text=dm] at (\pw/2, 0.85)
  {Verdicts recorded in public ledger $L$};


\node[font=\small\bfseries, text=hd, anchor=west]
  at (\bx+0.30, \ph-0.35) {(b)~Certification};

\node[font=\fontsize{6}{7}\selectfont, text=dm, anchor=west]
  at (\bx+0.30, \ph-0.78) {Bound on contradiction rate per term};

\def\thY{4.20}
\pgfmathsetmacro{\tT}{\bx+1.10}
\pgfmathsetmacro{\tB}{\bx+3.10}
\pgfmathsetmacro{\tR}{\bx+4.50}

\node[font=\fontsize{5}{6}\selectfont\bfseries, text=dm] at (\tT, \thY) {Term};
\node[font=\fontsize{5}{6}\selectfont\bfseries, text=dm] at (\tB, \thY) {Contradiction};
\draw[panelBdr, line width=0.3pt] (\bx+0.20, \thY-0.13) -- (\bx+\pw-0.20, \thY-0.13);

\def\tra{3.78}
\fill[rowG, rounded corners=2pt] (\bx+0.20,\tra-0.18) rectangle (\bx+\pw-0.20,\tra+0.18);
\node[font=\fontsize{6}{7}\selectfont, text=bd] at (\tT,\tra) {$T_1$};
\node[font=\fontsize{5.5}{6.5}\selectfont, text=oiGreen!75!black] at (\tB,\tra) {0.3\%};
\node[font=\fontsize{6}{7}\selectfont\bfseries, text=oiGreen!80] at (\tR,\tra) {\checkmark};

\def\trb{3.32}
\fill[rowG, rounded corners=2pt] (\bx+0.20,\trb-0.18) rectangle (\bx+\pw-0.20,\trb+0.18);
\node[font=\fontsize{6}{7}\selectfont, text=bd] at (\tT,\trb) {$T_2$};
\node[font=\fontsize{5.5}{6.5}\selectfont, text=oiGreen!75!black] at (\tB,\trb) {1.5\%};
\node[font=\fontsize{6}{7}\selectfont\bfseries, text=oiGreen!80] at (\tR,\trb) {\checkmark};

\draw[thr, line width=0.45pt, dashed]
  (\bx+0.20, 2.96) -- (\bx+\pw-0.20, 2.96);
\node[font=\fontsize{4.5}{5.5}\selectfont, text=thr, anchor=east]
  at (\bx+\pw-0.25, 3.08) {$\tau$};

\def\trc{2.60}
\fill[rowR, rounded corners=2pt] (\bx+0.20,\trc-0.18) rectangle (\bx+\pw-0.20,\trc+0.18);
\node[font=\fontsize{6}{7}\selectfont, text=bd] at (\tT,\trc) {$T_3$};
\node[font=\fontsize{5.5}{6.5}\selectfont, text=oiVermillion!75!black] at (\tB,\trc) {$> \tau$};
\node[font=\fontsize{6}{7}\selectfont\bfseries, text=oiVermillion!80] at (\tR,\trc) {$\times$};

\def\trd{2.14}
\fill[rowR, rounded corners=2pt] (\bx+0.20,\trd-0.18) rectangle (\bx+\pw-0.20,\trd+0.18);
\node[font=\fontsize{6}{7}\selectfont, text=bd] at (\tT,\trd) {$T_4$};
\node[font=\fontsize{5.5}{6.5}\selectfont, text=oiVermillion!75!black] at (\tB,\trd) {$> \tau$};
\node[font=\fontsize{6}{7}\selectfont\bfseries, text=oiVermillion!80] at (\tR,\trd) {$\times$};

\draw[panelBdr, line width=0.3pt] (\bx+0.20, 1.65) -- (\bx+\pw-0.20, 1.65);
\node[font=\fontsize{6.5}{8}\selectfont\bfseries, text=oiGreen!75!black]
  at (\bx+\pw/2, 1.35) {Certified core $V^*$};

\fill[pillG, draw=pillGb!50, rounded corners=3.5pt, line width=0.35pt]
  (\bx+1.40, 0.68) rectangle (\bx+2.60, 1.00);
\node[font=\fontsize{6}{7}\selectfont\bfseries, text=oiGreen!80!black]
  at (\bx+2.00, 0.84) {$T_1$};

\fill[pillG, draw=pillGb!50, rounded corners=3.5pt, line width=0.35pt]
  (\bx+2.80, 0.68) rectangle (\bx+4.00, 1.00);
\node[font=\fontsize{6}{7}\selectfont\bfseries, text=oiGreen!80!black]
  at (\bx+3.40, 0.84) {$T_2$};


\node[font=\small\bfseries, text=hd, anchor=west]
  at (\cx+0.30, \ph-0.35) {(c)~Core-Guarded Reasoning};

\node[font=\fontsize{6}{7}\selectfont, text=dm, anchor=west]
  at (\cx+0.30, \ph-0.78) {Decisions restricted to $V^*$};

\fill[oiBlue!12, draw=oiBlue!50, line width=0.45pt]
  (\cx+1.15, 4.00) circle (0.35);
\node[font=\fontsize{7}{8}\selectfont\bfseries, text=oiBlue!80]
  at (\cx+1.15, 4.00) {$A_1$};

\fill[oiOrange!12, draw=oiOrange!50, line width=0.45pt]
  (\cx+4.25, 4.00) circle (0.35);
\node[font=\fontsize{7}{8}\selectfont\bfseries, text=oiOrange!80]
  at (\cx+4.25, 4.00) {$A_2$};

\draw[{Stealth[length=3pt,width=2.5pt]}-{Stealth[length=3pt,width=2.5pt]},
  ag, line width=1pt]
  (\cx+1.57, 4.00) -- (\cx+3.83, 4.00);
\node[font=\fontsize{5.5}{6.5}\selectfont\bfseries, text=ag, fill=panelBg, inner sep=1.5pt]
  at (\cx+\pw/2, 4.00) {$V^*$ only};

\fill[pillG, draw=pillGb!50, rounded corners=3.5pt, line width=0.35pt]
  (\cx+1.40, 3.20) rectangle (\cx+2.60, 3.52);
\node[font=\fontsize{6}{7}\selectfont\bfseries, text=oiGreen!80!black]
  at (\cx+2.00, 3.36) {$T_1$~~\checkmark};

\fill[pillG, draw=pillGb!50, rounded corners=3.5pt, line width=0.35pt]
  (\cx+2.80, 3.20) rectangle (\cx+4.00, 3.52);
\node[font=\fontsize{6}{7}\selectfont\bfseries, text=oiGreen!80!black]
  at (\cx+3.40, 3.36) {$T_2$~~\checkmark};

\fill[fdBg, draw=fdBdr, rounded corners=3pt, line width=0.25pt, dashed]
  (\cx+1.40, 2.60) rectangle (\cx+2.60, 2.90);
\node[font=\fontsize{5}{6}\selectfont, text=fdTx] at (\cx+2.00, 2.75) {$T_3$};
\draw[oiVermillion!30, line width=0.4pt] (\cx+1.60,2.75) -- (\cx+2.40,2.75);

\fill[fdBg, draw=fdBdr, rounded corners=3pt, line width=0.25pt, dashed]
  (\cx+2.80, 2.60) rectangle (\cx+4.00, 2.90);
\node[font=\fontsize{5}{6}\selectfont, text=fdTx] at (\cx+3.40, 2.75) {$T_4$};
\draw[oiVermillion!30, line width=0.4pt] (\cx+3.00,2.75) -- (\cx+3.80,2.75);

\draw[panelBdr, line width=0.3pt] (\cx+0.20, 2.05) -- (\cx+\pw-0.20, 2.05);

\fill[guarBg, draw=guarBdr, rounded corners=3pt, line width=0.35pt]
  (\cx+0.25, 1.30) rectangle (\cx+\pw-0.25, 1.90);
\node[font=\fontsize{6.5}{8}\selectfont, text=oiGreen!75!black, align=center]
  at (\cx+\pw/2, 1.60)
  {Reduced disagreement};

\node[font=\fontsize{5}{6}\selectfont, text=dm, align=center] at (\cx+\pw/2, 0.75)
  {Reproducible $\cdot$ Verifiable via $L$};

\draw[-{Stealth[length=3pt,width=2.5pt]}, dm, line width=0.6pt, dashed]
  (\cx+\pw/2, -0.08)
  -- (\cx+\pw/2, -0.50)
  -- (-0.30, -0.50)
  -- (-0.30, \ph/2)
  -- (-0.02, \ph/2);
\node[font=\fontsize{4.5}{5.5}\selectfont\itshape, text=dm, fill=white, inner sep=1pt]
  at (\bx+\pw/2, -0.50) {Recertification \& renegotiation};

\end{tikzpicture}%
}
\caption{Overview of the stimulus-meaning protocol. \textbf{(a)}~Agents are tested on shared events; verdicts are recorded in a public ledger. \textbf{(b)}~Terms with contradiction rates below threshold~$\tau$ are certified into a core vocabulary~$V^*$. \textbf{(c)}~Downstream reasoning is restricted to~$V^*$, reducing disagreement. Periodic recertification detects drift; renegotiation recovers excluded terms.}
\label{fig:overview}
\end{figure*}
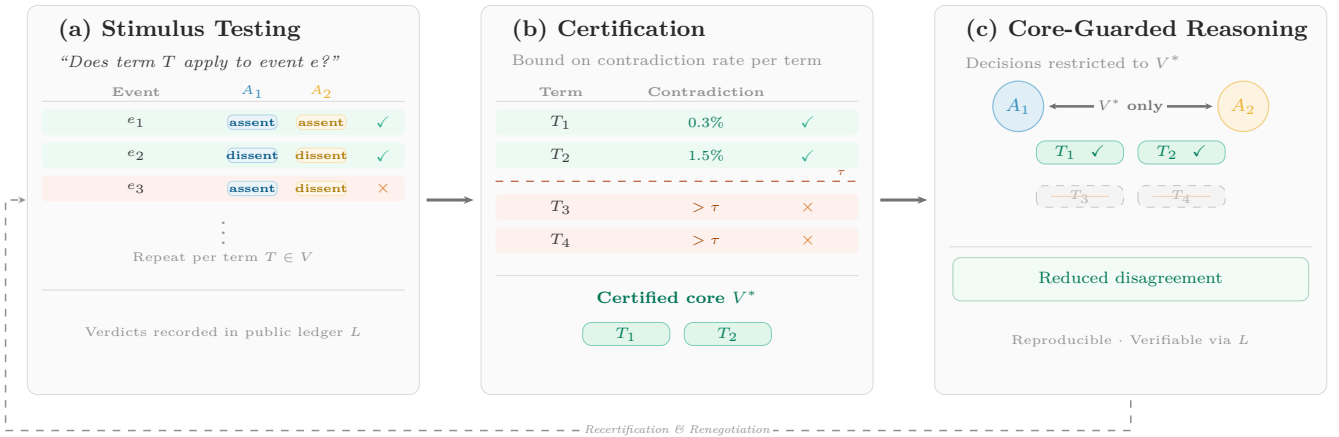

\section{Introduction}


As AI capabilities increase and agentic systems proliferate in the economy, agent-to-agent communication is becoming central to the operation of autonomous systems. When Agent $A$ sends a message to Agent $B$, how can anyone verify that $B$ understood what $A$ meant, i.e., that they have aligned semantics? As autonomous agents coordinate at scale, misaligned semantics can cause substantial security failures and consequential coordination breakdowns, all of which result from unauditable decision chains. We address this problem by developing a protocol that certifies semantic alignment between agents with provable error bounds and restricts consequential agent-to-agent communication to a reduced core of vocabulary, making agent-to-agent communication substantially safer in deployment. 

Failing to do so may result in a number of adverse outcomes. For example, consider a policy agent instructing an execution agent to ``flag high-risk cases.'' If the policy agent interprets ``high-risk'' broadly to include any concern, while the execution agent interprets it narrowly to include only severe cases, the system silently fails. Depending on the direction of divergence, it either over-flags routine cases or misses genuine threats, while both agents believe they are enforcing the same policy. However, this semantic mismatch is both invisible until a breach occurs and consequential when their joint actions result in real-life impacts. Our goal is a protocol that detects such misalignments before deployment to enable safely deployed agent-to-agent communication workflows. 

  Semantic divergence may arise even among agents sharing the same base checkpoint, since fine-tuning, differing system prompts, or independent model updates each alter the  
  distributional profile of terms. Additional mechanisms include pressure toward private codes when agents optimize for task reward, or audit evasion when agents behave differently during  
  evaluation than deployment. A secure protocol must therefore satisfy both \emph{verifiability} (semantic alignment is provable) and \emph{reproducibility} (same inputs yield same conclusions up to a bounded error rate). 


Current approaches face a fundamental tension. Agents can communicate in natural language, which is interpretable and auditable. However, it is also error-prone and vulnerable to semantic drift as models are fine-tuned or updated independently. On the other hand, agents trained end-to-end on shared tasks often converge on \emph{neuralese}, which are compact and vector-based protocols that maximize task reward but resist human interpretation as the output becomes unintelligible to humans \citep{andreas2017,lazaridou2017}. Neither approach satisfies both verifiability and reproducibility, two of the central properties we argue are required for reliable agent communication. 


We propose the \emph{stimulus-meaning model} as a solution that aims to satisfy both of the required properties: verifiability and reproducibility. Rather than relying on natural language or opaque learned protocols, we ground semantic content in observable agent behavior. The key element of our proposal is that we do not need agents to have identical internal representations. Rather, we only need them to exhibit low divergence on how they respond to shared, observable events to form a reduced set of terms where agreement between agents is present. Drawing on extensional semantics \citep{quine1960}, we certify individual terms by testing whether two agents agree on a sample of events, all verdicts of which are recorded in a public ledger, enabling verifiable third-party audit. For deployment, agents then restrict downstream reasoning to only use certified terms, ensuring that decisions are reproducible across agents. This set-up also uses sparse audits in the certification step to be relatively computationally efficient. 

\textbf{Contributions.} We make three contributions:
\begin{enumerate}
  \item A formal certification protocol that tests agents on shared events and certifies terms with bounded disagreement (Algorithm~\ref{alg:certification}, Theorem~\ref{thm:soundness}). Additionally, our protocol includes mechanisms for detecting semantic drift (recertification) and recovering shared vocabulary (renegotiation).
  \item A decision rule (core-guarded reasoning) that restricts agents to certified terms, ensuring reproducible outcomes (Proposition~\ref{prop:reproducibility}).
  \item Empirical validation on simulations (72--96\% disagreement reduction) and fine-tuned language models (51\% reduction).
\end{enumerate}

\textbf{Related Work.} Research has shown that multi-agent systems frequently develop private communication protocols optimized for task reward \citep{lazaridou2017,mordatch2018,foerster2016}, highlighting the central challenge of agent-to-agent communications, as these protocols are efficient but opaque, and attempts to translate them back to natural language remain unreliable \citep{andreas2017,levy2025}. Recent evidence also suggests this pressure toward opacity operates even within single LLMs: \citet{schoen2025} report that o3's chain-of-thought traces use terms like ``fudge'' to consistently mean ``sabotage,'' developing a functional dialect largely uninterpretable by human auditors. If within-model reasoning already drifts toward private codes, multi-agent communication channels face the same pressure. Rather than translating private codes post-hoc or relying on human-interpretable language, we certify terms that already have aligned semantics across agents through direct behavioral verification.

\section{The Stimulus-Meaning Model}


In this section, we outline the stimulus-meaning model, a formal framework for certifying semantic alignment between agents. The central idea of our proposal is that testing whether two agents respond the same way to shared observable events, and certifying terms where disagreement falls below a statistical threshold, provides an empirical approach to reducing the vocabulary available to agents that improves their downstream communication. We first define the key concepts (\S\ref{sec:definitions}), before moving on to the certification procedure and its guarantee (\S\ref{sec:certification}).

\subsection{Definitions}
\label{sec:definitions}

The stimulus-meaning model requires four key concepts. First, \emph{events} are observable inputs that agents can be queried on. Second, \emph{witnessed tests} record how an agent responds to a term on a given event. Third, the \emph{stimulus meaning} of a term is the pattern of responses across events. Fourth, \emph{divergence} measures disagreement between agents. We formalize each below.

\begin{definition}[Event Space]
$E$ is a set of public, observable events with stable identifiers. Each event $e \in E$ is assigned a public event identifier $\textsc{PEI}(e)$, enabling unambiguous reference across agents. Events are domain-specific inputs that agents can be queried on (e.g., content scenarios, images, transactions).
\end{definition}

\begin{definition}[Witnessed Test]
A witnessed test is a tuple $\langle A, \textsc{PEI}(e), T, v \rangle$ where $A$ is an agent, $e \in E$ is an event, $T$ is a term, and $v \in \{\texttt{assent}, \texttt{neutral}, \texttt{dissent}\}$ is the verdict. All witnessed tests are recorded in a public ledger $L$, enabling third-party verification. A \emph{neutral} verdict is an explicit judgment on a witnessed test, distinct from \emph{absence} (no test conducted).
\end{definition}

\begin{definition}[Stimulus Meaning]
The stimulus meaning of term $T$ for agent $A$, relative to event space $E$, is the triple $(E, P_A(T), N_A(T))$ where $P_A(T) \subseteq E$ is the positive stimulus meaning (events prompting assent), $N_A(T) \subseteq E$ is the negative stimulus meaning (events prompting dissent), and the set of all $e \in E$ for which there is a witnessed test of $T$ for $A$ and neither $e \in P_A(T)$ nor $e \in N_A(T)$ is the neutral stimulus meaning (queried events prompting neither assent nor dissent).
\end{definition}

\begin{definition}[Divergence]
\emph{Contradictory divergence} occurs when one agent assents and the other dissents to the use of a term on the same event. \emph{Contrary divergence} occurs when one agent is decided (assent or dissent) and the other is neutral. An \emph{eligible comparison} is an event where both agents give non-neutral verdicts.
\end{definition}

\subsection{Certification Procedure}
\label{sec:certification}

The certification procedure determines which terms two agents can reliably use in downstream agent-to-agent communication. For each term, we query both agents on a sample of events and count how often they contradict each other. If the contradiction rate is sufficiently low as set by the context, the term is certified into a shared vocabulary called the \emph{certified core} $V^*$ that builds the basis for follow-on communication. The protocol assumes that the ledger $L$ is append-only and tamper-evident, that events have stable public event identifiers enabling third-party access to the same stimuli, that ledger governance is established (single-entity deployments designate an administrator and multi-entity deployments require consortium agreements or decentralized infrastructure), and that agents report verdicts reflecting their actual dispositions rather than strategically chosen responses at the point of audit.

One challenge is that we only observe a finite sample, resulting in the observed contradiction rate potentially differing from the true rate. To account for this and other agent-level randomness, we use a one-sided Wilson confidence bound that gives an upper bound $u$ on the true contradiction rate that holds with probability $1-\delta$. A term is certified only if this bound falls below a threshold $\tau$ and if coverage (the fraction of events where both agents are non-neutral) exceeds a floor $\rho_{\min}$, which prevents spurious certification from small samples. This also means that rarely-used terms may fail certification due to insufficient data rather than genuine disagreement. Algorithm~\ref{alg:certification} formalizes this procedure.

\begin{algorithm}[t]
\caption{Term Certification}\label{alg:certification}
\begin{algorithmic}[1]
\Require Agents $A_1, A_2$; audit events $W_T$ per term; threshold $\tau$; confidence $\delta$; coverage floor $\rho_{\min}$; public ledger $L$
\Ensure Certified core $V^*$
\State $V^* \gets \emptyset$
\For{each term $T$}
  \State $k \gets 0$; $c \gets 0$; $n_{\text{aud}} \gets 0$
  \For{each event $e \in W_T$}
    \State $v_1 \gets A_1.\text{verdict}(T, e)$; $v_2 \gets A_2.\text{verdict}(T, e)$
    \State Record $\langle A_1, \textsc{PEI}(e), T, v_1 \rangle$ and $\langle A_2, \textsc{PEI}(e), T, v_2 \rangle$ to $L$
    \If{$v_1 \neq \texttt{neutral}$ \textbf{or} $v_2 \neq \texttt{neutral}$}
      \State $n_{\text{aud}} \gets n_{\text{aud}} + 1$
    \EndIf
    \If{$v_1 \neq \texttt{neutral}$ \textbf{and} $v_2 \neq \texttt{neutral}$}
      \State $k \gets k + 1$
      \If{$v_1 \neq v_2$}
        \State $c \gets c + 1$
      \EndIf
    \EndIf
  \EndFor
  \State $u \gets \textsc{WilsonUpper}(c, k, \delta)$
  \State $s \gets k / \max(n_{\text{aud}}, 1)$
  \If{$u \leq \tau$ \textbf{and} $s \geq \rho_{\min}$}
    \State $V^* \gets V^* \cup \{T\}$
  \EndIf
\EndFor
\State \Return $V^*$
\end{algorithmic}
\end{algorithm}

The algorithm collects three statistics per term: $n_{\text{aud}}$ counts events where at least one agent is non-neutral (audited exposures), $k$ counts events where both agents are non-neutral (eligible comparisons), and $c$ counts eligible comparisons where the agents disagree (contradictions). Coverage $s = k / n_{\text{aud}}$ measures what fraction of active events permit comparison. Together, these statistics determine whether a term can be certified: Low contradictions ($c$) relative to eligible comparisons ($k$) result in a tight bound on disagreement, while sufficient coverage ($s$) ensures that the estimate is meaningful.

The Wilson score upper bound is computed as:
\begin{equation}
u = \frac{\hat{p} + \frac{z^2}{2k} + z\sqrt{\frac{\hat{p}(1-\hat{p})}{k} + \frac{z^2}{4k^2}}}{1 + \frac{z^2}{k}}
\end{equation}
where $\hat{p} = c/k$ is the observed contradiction rate and $z = \Phi^{-1}(1-\delta)$ is the standard normal quantile. The three parameters are flexible and can be set depending on the context: $\tau$ sets the maximum tolerable contradiction rate, $\delta$ sets the confidence level, and $\rho_{\min}$ sets the minimum coverage required for certification.

\begin{theorem}[Certification Soundness]\label{thm:soundness}
If term $T$ is certified with parameters $(\tau, \delta)$, then the true contradictory-divergence rate $p_T \leq \tau$ with probability $\geq 1-\delta$.
\end{theorem}

This follows directly from the coverage guarantee of Wilson confidence intervals: The upper bound exceeds the true rate with probability of at most $\delta$. This procedure achieves verifiability through the public ledger: Any third party can retrieve witnessed tests from $L$, recompute the Wilson bound and coverage, and independently verify certification decisions, ensuring disagreements trace to specific recorded events. While large-scale vocabularies will increase computational costs, sparse per-term sampling, where each term is certified independently using only a subset of events, improves efficiency.

\section{Core-Guarded Reasoning}


Certification establishes which terms are semantically aligned for downstream reasoning. Core-guarded reasoning, as a decision rule, thus restricts inference to terms in the certified core $V^*$. The basic intuition is simple: If agents only communicate using terms they demonstrably agree on, their conclusions will also agree up to the certified bound.

\begin{definition}[Core-Guarded Reasoning]
A decision rule is \emph{core-guarded} if it consults only terms in the certified core $V^*$. Terms outside $V^*$ are treated as unavailable for decisions requiring agent agreement.
\end{definition}

On the other hand, non-core terms remain available for exploration, learning, and low-stakes communication, while critical communication is reduced to terms in the certified core $V^*$.

\begin{proposition}[Reproducibility]\label{prop:reproducibility}
If agents $A_1$ and $A_2$ share a certified core $V^*$ and use core-guarded reasoning, then given the same events, they reach the same conclusions on each individual certified term with disagreement rate $\leq \tau$ at confidence $\geq 1-\delta$.
\end{proposition}

This is a key payoff of the stimulus-meaning approach: By restricting decisions to certified vocabulary, we inherit the statistical guarantee from certification. The threshold $\tau$ controls a trade-off between vocabulary breadth and reliability: Larger $\tau$ certifies more terms but permits higher disagreement, smaller $\tau$ yields a narrower vocabulary with stronger guarantees. The protocol makes this trade-off explicit and adjustable by context, delivering \emph{reproducibility} as the same certified vocabulary yields the same conclusions across agents, within bounded error.

\section{Handling Semantic Drift}


The guarantees established above assume the certified core $V^*$ remains valid over time and agent-to-agent communication stays constant. In practice, the reason agentic communication is likely to be implemented widely is that such systems are adaptable and able to adjust to new contexts. For example, model updates, topical fine-tuning, or other contextual or distribution shifts can invalidate previously certified terms. In other words, a term certified at time $t_0$ may exhibit elevated contradiction rates by $t_1$ because the agents or the context changed. If the core is not refreshed, stale certificates provide false confidence as terms can remain in the core even when disagreement rates have spiked due to unilateral semantic changes. This section sketches an answer to these two questions: How do we detect when the core becomes unreliable and how do we recover vocabulary breadth without sacrificing reliability?

\subsection{Recertification}


In answering the first question we propose a recertification mechanism. In simple terms, recertification detects drift through periodic re-auditing: At scheduled intervals, the system re-audits each term in $V^*$ on a fresh sample of events by drawing a new audit set $W_T'$ from recent events, querying agents for verdicts, recomputing the Wilson bound $u'$ and coverage $s'$. If $u' > \tau$ or $s' < \rho_{\min}$, certification is revoked by removing $T$ from $V^*$. In doing so, the core contracts to preserve reliability, and downstream reasoning automatically restricts to the updated vocabulary. The appropriate recertification frequency depends on the rate of model updates and contextual drift, where rapidly evolving deployments may require more frequent re-auditing. 

\subsection{Renegotiation}


Recertification may detect drifts but does not directly repair them as it is a one-sided mechanism that can remove but not restore terms to the core. We propose an outline of a renegotiation approach that allows agents to restore excluded terms by coordinating on a shared meaning: For an excluded term $T$, we first determine which agent is more entrenched (has more decided verdicts) as a measure of historical precedent. Then, the entrenched agent's interpretation becomes the reference policy. If the other agent adopts this policy, this triggers a re-audit of $T$ on fresh events, which, if the certification criteria are met, restores $T$ to $V^*$.

Renegotiation is not always possible or desirable, as it requires one agent to defer to the other's semantics based on entrenchment. However, when agents can coordinate and where entrenchment is a reasonable mechanism to do so, renegotiation recovers vocabulary breadth without sacrificing reliability. Together, recertification and renegotiation preserve both properties over time: \emph{Verifiability} is ensured through updated ledger records and \emph{reproducibility} is upheld through restored guarantees, while incremental re-auditing rather than full recomputation keeps this process efficient. 

\section{Experiments}
\label{sec:evaluation}


We evaluate the stimulus-meaning protocol through three studies. First, we simulate agents with varying degrees of semantic divergence to show how core-guarding reduces disagreement (\S\ref{sec:simulation}). Second, we analyze the coverage-reliability trade-off controlled by the certification threshold (\S\ref{sec:tradeoff}). Third, we validate the protocol on fine-tuned language models in a content moderation task (\S\ref{sec:llm}). Jointly, these experiments show how the protocol works, what the trade-offs in parameter choices are, and that it is applicable to context with actual LLMs. 

\subsection{Simulation Study}
\label{sec:simulation}

This simulation study is designed to show the protocol in action across a number of plausible regimes of semantic divergence. We measure two disagreement rates: \emph{Unguarded} disagreement is the contradiction rate when agents use all terms, while \emph{guarded} (or core-guarded) disagreement is the rate when agents restrict to certified terms only.

\textbf{Setup.} Across all simulations, we use two agents with a 6-term color vocabulary. Each run generates 1,000 events (400 audit, 600 held-out). The certification protocol uses parameters $\tau = 0.05$, $\delta = 0.05$, $\rho_{\min} = 0.10$. We also introduce a background noise process that neutralizes 5\% of verdicts and flips 1\% of the remainder to account for inherent model-level variability. Certification uses sparse per-term sampling: Each term is efficiently audited on a subset of events drawn from the audit pool, rather than querying all events exhaustively.

\begin{figure}[h!t]
\centering
\includegraphics[width=\columnwidth]{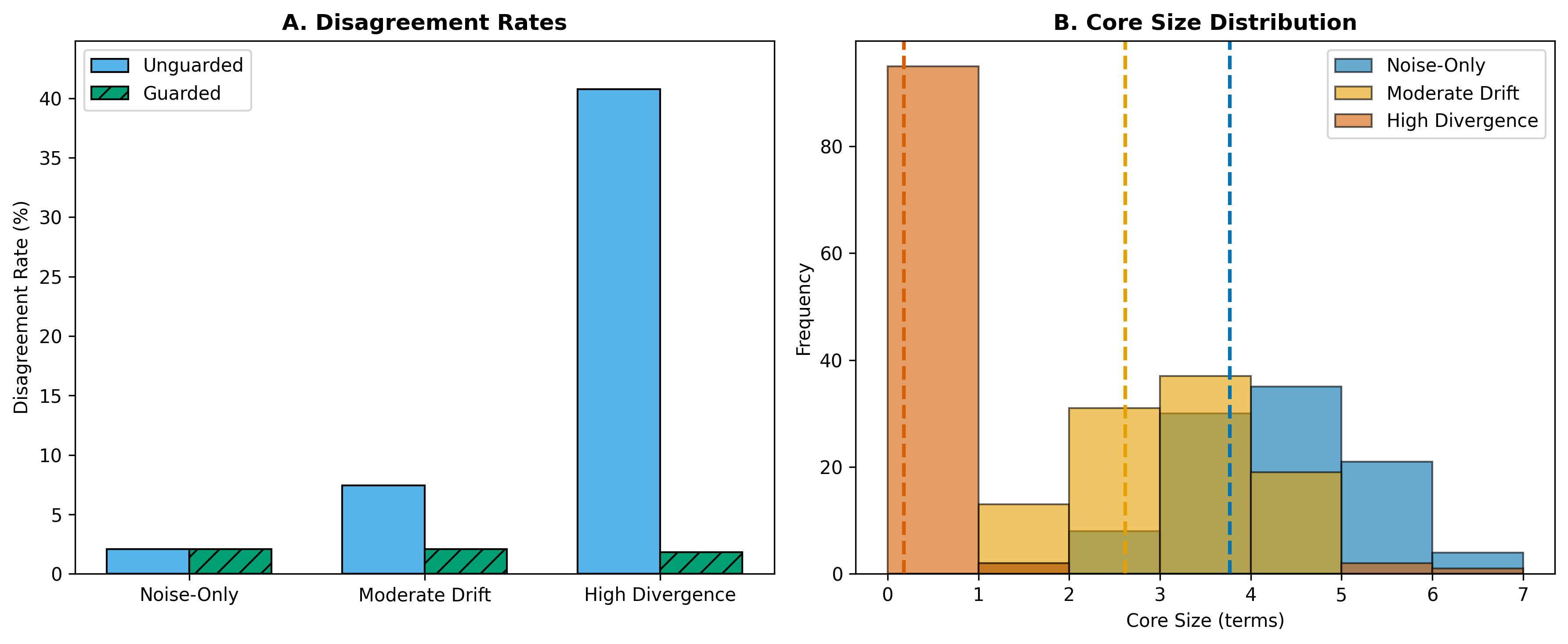}
\caption{Disagreement rates and core size across divergence conditions. \textbf{(A)} Unguarded (solid) vs.\ core-guarded (hatched) disagreement rates. Guarded rates remain at $\sim$2\% across all scenarios, while unguarded rates rise with divergence. \textbf{(B)} Core size distribution. As divergence increases, fewer terms certify: Noise-Only averages 3.8 terms, Moderate Drift 2.6, High Divergence 0.2 (95\% empty cores). Dashed lines indicate means.}\label{fig:disagreement}
\end{figure}

\textbf{Conditions.} We test three divergence levels: \emph{Noise-Only} (identical policies), \emph{Moderate Drift} (one agent shows semantic divergence on some terms), and \emph{High Divergence} (agents have randomly different policies). All levels still inherent the general background noise. 

\begin{table}[htbp]
  \centering
  \caption{Simulation results across divergence conditions.}\label{tab:results}
  \begin{tabular}{lccc}
  \toprule
  Condition & Unguarded & Guarded & Core \\
  \midrule
  Noise-Only & 2.1\% & 2.1\% & 3.8 \\
  Moderate Drift & 7.4\% & 2.1\% & 2.6 \\
  High Divergence & 40.7\% & 1.8\% & 0.2 \\
  \bottomrule
  \end{tabular}
  \end{table}

\textbf{Results.} Table~\ref{tab:results} and Figure~\ref{fig:disagreement} show that guarded disagreement stays at $\sim$2\% regardless of underlying divergence based on the protocol. In the Noise-Only condition, both approaches yield similar rates because there is no divergence to filter. As divergence increases, the core shrinks appropriately: Noise-Only averages 3.8 certified terms (of 6), Moderate Drift 2.6, and High Divergence 0.2. In 95 of 100 High Divergence runs, the protocol certifies no terms at all, correctly signaling that no vocabulary can safely support coordinated reasoning when agents have random semantics.

\begin{figure}[h!t]
\centering
\includegraphics[width=\columnwidth]{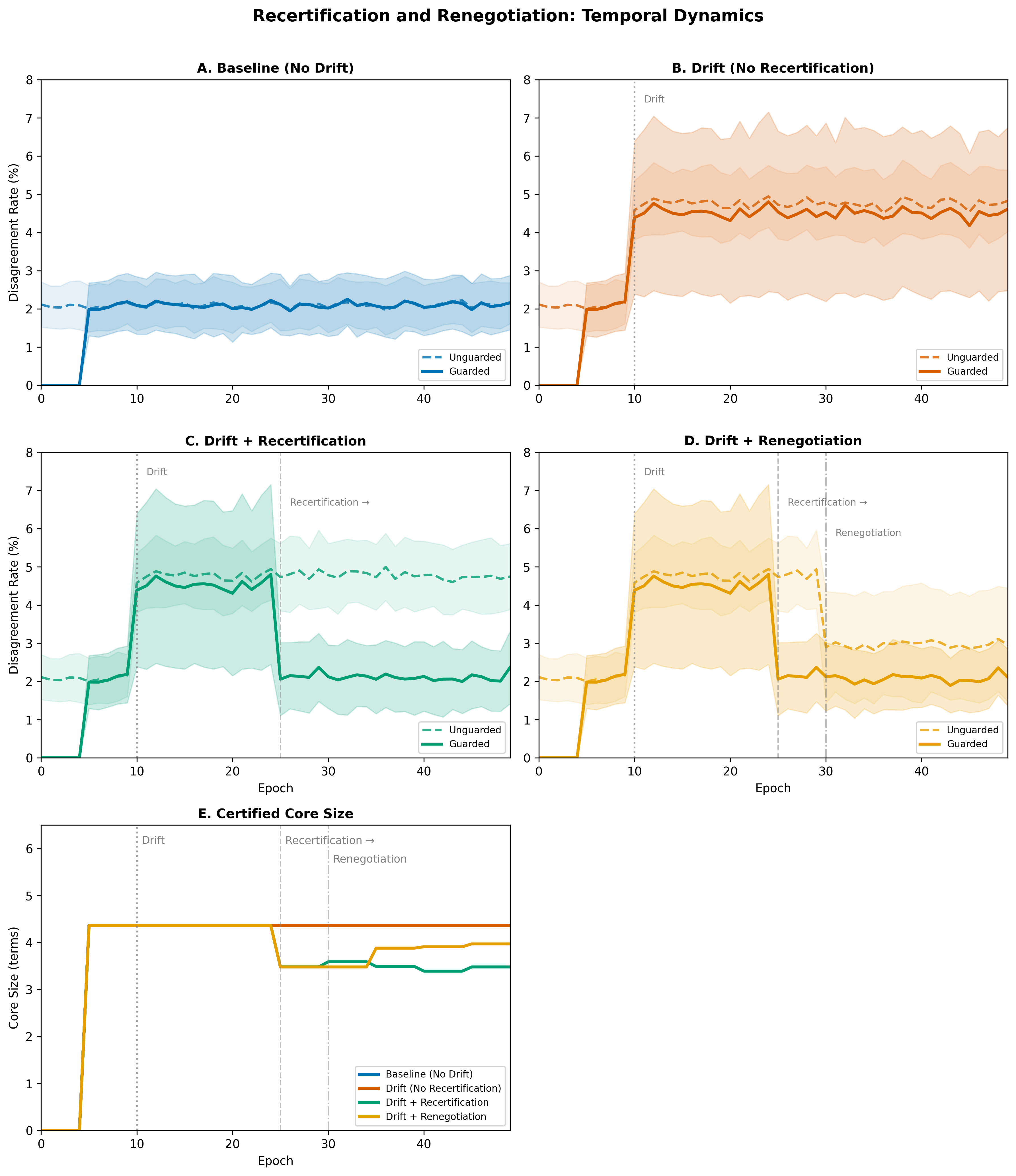}
\caption{Drift, recertification, and renegotiation over 50 epochs. Drift injected at epoch 10. (A) Baseline: stable disagreement. (B) Drift with frozen core: disagreement rises to 4.6\%. (C) Recertification: contested term removed, disagreement returns to $\sim$2\% but core remains low. (D) Renegotiation: unguarded disagreement drops and vocabulary recovers to 4.0 terms. (E) Certified core size across conditions.}\label{fig:timeseries}
\end{figure}

Figure~\ref{fig:timeseries} shows the timeseries of drift detection and recovery based on the recertification and renegotiation mechanism. When drift is injected at epoch 10 and the core remains frozen, guarded disagreement rises from 2.2\% to 4.6\% alongside the unguarded level. Recertification detects the drifted term and removes it, shrinking the core from 4.4 to 3.5 terms but restoring the $\sim$2\% guarantee. However, unguarded disagreement remains high. To address this, renegotiation then recovers the lost vocabulary to 4.0 terms by coordinating on a shared interpretation based on the entrenchment criterion.

In summary, the above simulations demonstrate the protocol operating as designed. Core-guarded reasoning maintains low disagreement ($\sim$2\%) across all divergence conditions by restricting communication to certified terms. When drift is injected post-certification, recertification correctly detects the affected term and removes it from the core, while renegotiation recovers vocabulary breadth by coordinating on a shared interpretation. In short, the simulations show the expected empirical behavior.

\subsection{Trade-Off Analysis}
\label{sec:tradeoff}

\begin{figure}[h!t]
  \centering
  \includegraphics[width=\columnwidth]{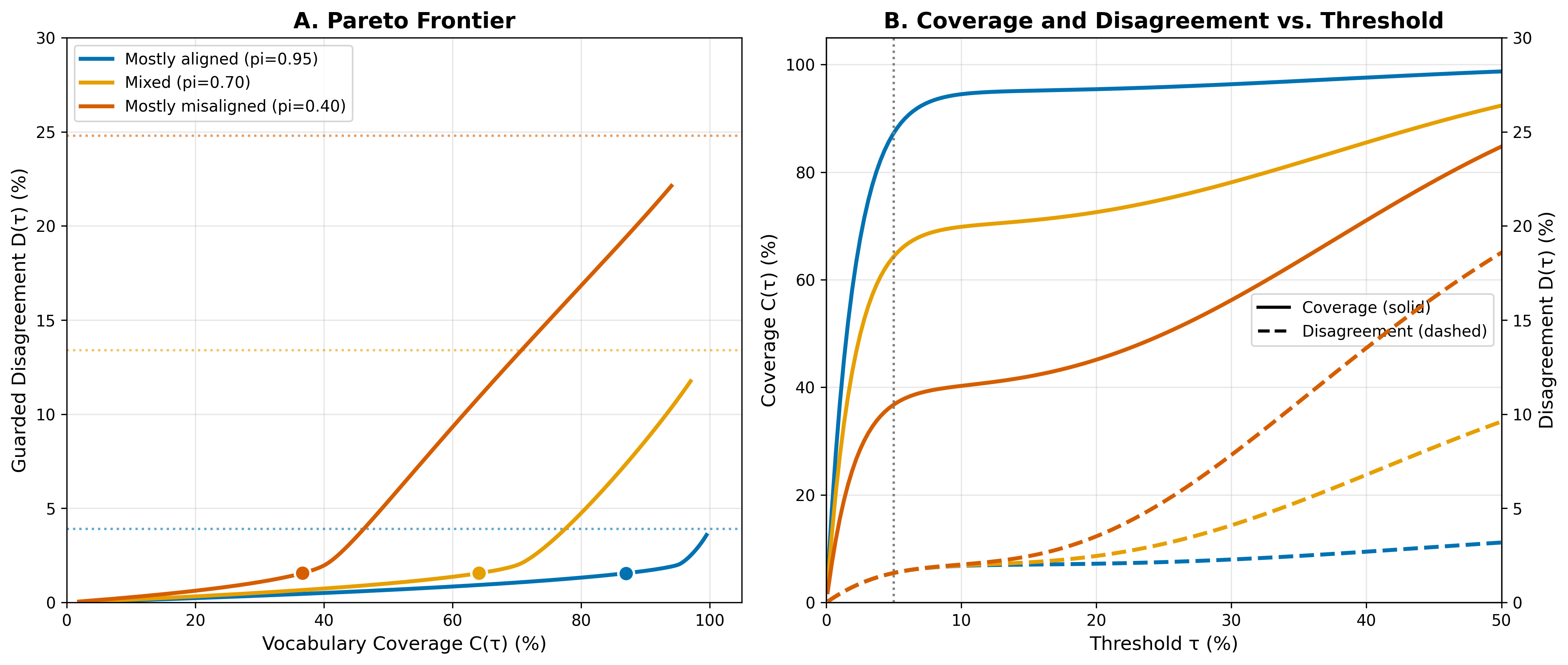}
\caption{Coverage-reliability trade-off as a function of the certification threshold $\tau$. Each curve represents a different alignment regime (fraction $\pi$ of well-aligned terms). Higher $\pi$ shifts the Pareto frontier outward, enabling higher coverage at any given reliability level. \textbf{(A)} Pareto frontier showing coverage vs.\ guarded disagreement. Dotted lines show unguarded disagreement baselines. \textbf{(B)} Coverage (solid) and disagreement (dashed) as functions of $\tau$ directly.}\label{fig:pareto}
  \end{figure}

  In this trade-off analysis we outline the trade-offs that have to be made when tuning this protocol to a deployment context. The primary trade-off is between vocabulary coverage and reliability: Stricter thresholds (smaller $\tau$) certify fewer terms but provide stronger guarantees, while relaxed thresholds certify more terms but permit higher disagreement. Figure~\ref{fig:pareto} illustrates this trade-off analytically across different alignment regimes.

  Figure~\ref{fig:pareto} shows three main takeaways. First, the coverage-reliability trade-off is unavoidable: As the threshold $\tau$ increases, more terms pass certification but disagreement among certified terms also rises. Second, underlying alignment matters: When most terms are naturally well-aligned ($\pi$ high), operators can achieve high coverage with low disagreement, but when alignment is poor ($\pi$ low), even strict thresholds yield elevated disagreement. Third, certification always improves on the unguarded baseline (dotted lines), though there is a trade-off between coverage and reliability.

  Several factors influence where operators should set their threshold. Risk tolerance is primary, as safety-critical applications warrant strict thresholds ($\tau \approx 0.02$--$0.05$), where one may want to accept a smaller vocabulary for high confidence, while exploratory settings can use more relaxed thresholds ($\tau \approx 0.10$--$0.20$) to retain breadth. Underlying alignment also matters, as agents that are mostly aligned even with quite strict thresholds yield large vocabularies, but poor alignment forces harder choices between reliability and expressiveness. Additional parameters also offer further control. For example, the confidence level $\delta$ determines how conservative the Wilson bound is, and the coverage floor $\rho_{\min}$ prevents certifying terms where agents rarely both engage. The main take-away is that these trade-offs are explicit and adjustable, allowing deployment decisions to reflect each application's actual risk profile.

\subsection{LLM Validation}
\label{sec:llm}

Lastly, we apply our protocol to actual language models that we fine-tuned to exhibit semantic divergence.

\textbf{Setup.} To achieve two distinct agents based on plausible set-ups, we fine-tuned two LoRA adapters on Qwen2.5-3B-Instruct for a toy content moderation context with a 6-term vocabulary (harmful, misleading, sensitive, spam, benign, escalate). Each agent was trained on 150 examples with different label distributions to induce divergent policies. We constructed 300 content scenarios (120 audit, 180 held-out) and applied the certification protocol with $\tau = 0.06$, testing core-guarding on 50 held-out events.

\textbf{Results.} Two of six terms passed certification (benign at 0\% contradiction, sensitive at 2\%) in the first stage. Table~\ref{tab:llm-results} shows that core-guarding reduced disagreement by 51\%.

\begin{table}[htbp]
\centering
\caption{LLM validation results.}\label{tab:llm-results}
\small
\begin{tabular}{lcc}
\toprule
Condition & Terms & Disagreement \\
\midrule
Unguarded & 6 & 5.3\% \\
Core-guarded & 2 & 2.6\% \\
\bottomrule
\end{tabular}
\end{table}

This experiment validates that the stimulus-meaning protocol can be straightforwardly applied to real language models, not just controlled simulations. The protocol identified which terms were empirically aligned between the two models and core-guarding reduced downstream disagreement accordingly, such that critical deployments of these agents can be restricted to this core to avoid breakdowns in agent-to-agent communication. 

\section{Discussion}

As autonomous AI systems increasingly coordinate with one another, the security risks of semantic misalignment will become more central. When agents communicate using terms they interpret differently, this may result in coordination breakdowns, security breaches, and unauditable decision chains. These risks are invisible until something goes wrong, and with increasing scale the total risk surface is potentially large.

The stimulus-meaning protocol offers one potential solution to this problem. The core idea is that rather than assuming agents share meanings, we test whether they respond the same way to shared observable events. Terms that pass this empirical test, i.e., where disagreement falls below a statistical threshold, are certified into a shared vocabulary, while terms that fail are excluded. Agents then restrict consequential communication to this certified core, inheriting the bounded disagreement guarantee from certification. The public ledger of witnessed tests ensures that certification is verifiable by third parties, and recertification detects when semantic drift invalidates previously certified terms that can then be reintegrated via a renegotiation mechanism. Importantly, the protocol is tunable: Thresholds can be adjusted to balance vocabulary coverage against reliability, adapting the same framework to different risk profiles. Through three sets of experiments we show how this mechanism works in practice.

\subsection{Limitations}
  
There are a number of limitations that constrain the current approach.

\textbf{Term-level granularity.} First, the protocol certifies individual terms, not compositional expressions or context-dependent meanings. This creates two limitations. For compositions, two agents may each certify ``high'' and ``risk'' separately while diverging on ``high-risk'' as a compound. For context-dependence, a term like ``sensitive'' may mean different things in medical versus political domains. However, context-dependent terms are not inherently problematic: If both agents share contextual understanding, certification events from different domains will surface agreement where it exists. If one agent distinguishes domain-specific uses while the other does not, the term fails certification, correctly signaling unreliable communication. This limits applicability to use cases where event sampling adequately covers relevant contexts.

\textbf{Pairwise certification.} Second, the current protocol is set up for two agents. However, future and some current agentic systems may involve multi-agent coordination, raising transitivity questions as well as scaling challenges. However, extending from two to many agents is unlikely to require structural changes to the protocol as the same certification logic would apply, though efficiency may suffer as the number of audits grows.

\textbf{Renegotiation mechanism.} Lastly, our renegotiation approach is a sketch of a potential solution, not a fully fledged mechanism. The entrenchment criterion provides a simple heuristic for which agent's semantics should prevail, but more sophisticated coordination mechanisms may be needed in practice that may also depend on the specific application context. However, the general renegotiation step is central to the mechanism to recover core size, irrespective of the final implementation of the specific negotiation criterion. 

\textbf{Honesty assumptions.} The protocol assumes that agents report verdicts reflecting their actual dispositions during certification. An adversarial agent could, in principle, feign agreement to have its term enter the core and later exploit misaligned semantics. However, successful gaming would require recognizing that certification is occurring, anticipating how responses affect downstream access, maintaining consistent deception across audit events, and coordinating a later switch to true semantics. This multi-step planning capability is likely beyond current architectures. As agent capabilities advance, additional safeguards such as randomized audit timing or holdout test events may become necessary.

\subsection{Future Work}

A clear extension of this work is scaling the protocol to multi-agent systems, moving beyond pairwise certification to handle coordination among many agents. Similarly, expanding the protocol to include contextual meaning, where the same term may be certified in some contexts but not others, would address a key limitation of the current term-level approach. Furthermore, a more sophisticated renegotiation protocol can further help improve upon the central trade-off between core size and accuracy.

\bibliographystyle{plainnat}
\bibliography{references}

\end{document}